\documentclass[10pt,logo,copyright]{xhumanoid-report}
\usepackage[authoryear,sort&compress,round]{natbib}

\usepackage[utf8]{inputenc} %
\usepackage[T1]{fontenc}    %

\usepackage{parskip}        %
\usepackage{url}            %
\usepackage{booktabs}       %
\usepackage{amsfonts}       %
\usepackage{nicefrac}       %
\usepackage{microtype}      %
\usepackage{xcolor}         %
\usepackage[dvipsnames]{xcolor} %
\usepackage{graphicx}
\usepackage{animate}        %
\usepackage{subcaption}
\usepackage{tabularx}
\usepackage{makecell}
\usepackage{adjustbox}
\usepackage{setspace}
\usepackage{todonotes}
\usepackage{colortbl}
\usepackage{wrapfig}

\newcolumntype{M}[1]{>{\centering\arraybackslash}m{#1}}
\usepackage{float}
\usepackage{tikz}
\usetikzlibrary{positioning,shapes,arrows}
\usepackage{amsmath,amsfonts,bm, bbm,leftindex}
\usepackage{multirow}
\usepackage{comment}
\usepackage{gensymb}
\usepackage{lipsum}
\usetikzlibrary{arrows.meta, positioning, fit}
\usepackage[para]{threeparttable}
\usepackage{tikz}
\usetikzlibrary{tikzmark}

\def\eqref#1{equation~\ref{#1}}

\def\1{\bm{1}}

\DeclareMathAlphabet{\mathsfit}{\encodingdefault}{\sfdefault}{m}{sl}
\SetMathAlphabet{\mathsfit}{bold}{\encodingdefault}{\sfdefault}{bx}{n}

\makeatletter
\let\save@mathaccent\mathaccent
\newcommand*\if@single[3]{%
  \setbox0\hbox{${\mathaccent"0362{#1}}^H$}%
  \setbox2\hbox{${\mathaccent"0362{\kern0pt#1}}^H$}%
  \ifdim\ht0=\ht2 #3\else #2\fi
  }
\newcommand*\rel@kern[1]{\kern#1\dimexpr\macc@kerna}
\newcommand*\widebar[1]{\@ifnextchar^{{\wide@bar{#1}{0}}}{\wide@bar{#1}{1}}}
\newcommand*\wide@bar[2]{\if@single{#1}{\wide@bar@{#1}{#2}{1}}{\wide@bar@{#1}{#2}{2}}}
\newcommand*\wide@bar@[3]{%
  \begingroup
  \def\mathaccent##1##2{%
    \let\mathaccent\save@mathaccent
    \if#32 \let\macc@nucleus\first@char \fi
    \setbox\z@\hbox{$\macc@style{\macc@nucleus}_{}$}%
    \setbox\tw@\hbox{$\macc@style{\macc@nucleus}{}_{}$}%
    \dimen@\wd\tw@
    \advance\dimen@-\wd\z@
    \divide\dimen@ 3
    \@tempdima\wd\tw@
    \advance\@tempdima-\scriptspace
    \divide\@tempdima 10
    \advance\dimen@-\@tempdima
    \ifdim\dimen@>\z@ \dimen@0pt\fi
    \rel@kern{0.6}\kern-\dimen@
    \if#31
      \overline{\rel@kern{-0.6}\kern\dimen@\macc@nucleus\rel@kern{0.4}\kern\dimen@}%
      \advance\dimen@0.4\dimexpr\macc@kerna
      \let\final@kern#2%
      \ifdim\dimen@<\z@ \let\final@kern1\fi
      \if\final@kern1 \kern-\dimen@\fi
    \else
      \overline{\rel@kern{-0.6}\kern\dimen@#1}%
    \fi
  }%
  \macc@depth\@ne
  \let\math@bgroup\@empty \let\math@egroup\macc@set@skewchar
  \mathsurround\z@ \frozen@everymath{\mathgroup\macc@group\relax}%
  \macc@set@skewchar\relax
  \let\mathaccentV\macc@nested@a
  \if#31
    \macc@nested@a\relax111{#1}%
  \else
    \def\gobble@till@marker##1\endmarker{}%
    \futurelet\first@char\gobble@till@marker#1\endmarker
    \ifcat\noexpand\first@char A\else
      \def\first@char{}%
    \fi
    \macc@nested@a\relax111{\first@char}%
  \fi
  \endgroup
}
\makeatother

\definecolor{darkred}{rgb}{0.7, 0.0, 0.0}

\usepackage{amsmath} 

\usepackage{pifont}

\usepackage[nameinlink]{cleveref}
\crefname{equation}{Eq.}{Eqs.}
\crefname{figure}{Fig.}{Figs.}
\crefname{section}{Sec.}{Sec.}
\crefname{appendix}{App.}{App.}
\crefname{table}{Tab.}{Tabs.}
\crefname{algorithm}{Algo}{Algo}
\crefname{thm}{Thm}{Thm}
\Crefname{thm}{Thm}{Thm}
\crefname{prop}{Prop}{Prop}

\newcommand{\crefnames}[3]{%
  \@for\next:=#1\do{%
    \expandafter\crefname\expandafter{\next}{#2}{#3}%
  }%
}

\title{MeshMimic: Geometry-Aware Humanoid Motion Learning through 3D Scene Reconstruction}

\author{
    \textbf{Qiang Zhang}$^{1,2,*\dagger}$, \textbf{Jiahao Ma}$^{1,7,*}$, \textbf{Peiran Liu}$^{1,2,*}$, \textbf{Shuai Shi}$^{1,*}$, \textbf{Zeran Su}$^{1}$, \textbf{Zifan Wang}$^{2}$, \textbf{Jingkai Sun}$^{1,3}$, \textbf{Wei Cui}$^{1}$, 
    \textbf{Jialin Yu}$^{1}$, \textbf{Gang Han}$^{1}$, \textbf{Wen Zhao}$^{1}$, \textbf{Pihai Sun}$^{1}$, \textbf{Kangning Yin}$^{6}$, \textbf{Jiaxu Wang}$^{5}$, \textbf{Jiahang Cao}$^{3}$, \textbf{Lingfeng Zhang}$^{4}$, \textbf{Hao Cheng}$^{2}$, 
    \textbf{Xiaoshuai Hao}$^{4}$, \textbf{Yiding Ji}$^{2}$, \textbf{Junwei Liang}$^{2}$, \textbf{Jian Tang}$^{1}$, \textbf{Renjing Xu}$^{2}$, \textbf{Yijie Guo}$^{1}$ \\
    
    \small
    $^1$X-Humanoid \\
    $^2$The Hong Kong University of Science and Technology (Guangzhou) \quad
    $^3$The University of Hong Kong \quad
    $^4$Tsinghua University \\
    $^5$The Chinese University of Hong Kong \quad
    $^6$Shanghai Jiao Tong University \quad
    $^7$The Australian National University \\
    
    \footnotesize
    $^*$Equal Contribution \quad $^\dagger$Corresponding Author
}

\begin{abstract}
Humanoid motion control has witnessed significant breakthroughs in recent years, with deep reinforcement learning (RL) emerging as a primary catalyst for achieving complex, human-like behaviors. However, the high dimensionality and intricate dynamics of humanoid robots make manual motion design impractical, leading to a heavy reliance on expensive motion capture (MoCap) data. These datasets are not only costly to acquire but also frequently lack the necessary geometric context of the surrounding physical environment. Consequently, existing motion synthesis frameworks often suffer from a decoupling of motion and scene, resulting in physical inconsistencies such as contact slippage or mesh penetration during terrain-aware tasks.

In this work, we present MeshMimic, an innovative framework that bridges 3D scene reconstruction and embodied intelligence to enable humanoid robots to learn coupled "motion-terrain" interactions directly from video. By leveraging state-of-the-art 3D vision models, our framework precisely segments and reconstructs both human trajectories and the underlying 3D geometry of terrains and objects. We introduce an optimization algorithm based on kinematic consistency to extract high-quality motion data from noisy visual reconstructions, alongside a contact-invariant retargeting method that transfers human-environment interaction features to the humanoid agent. Experimental results demonstrate that MeshMimic achieves robust, highly dynamic performance across diverse and challenging terrains. Our approach proves that a low-cost pipeline utilizing only consumer-grade monocular sensors can facilitate the training of complex physical interactions, offering a scalable path toward the autonomous evolution of humanoid robots in unstructured environments.
\end{abstract}

\begin{document}

\maketitle

\begin{center}
  \captionsetup{type=figure}
  \includegraphics[width=0.75\textwidth]{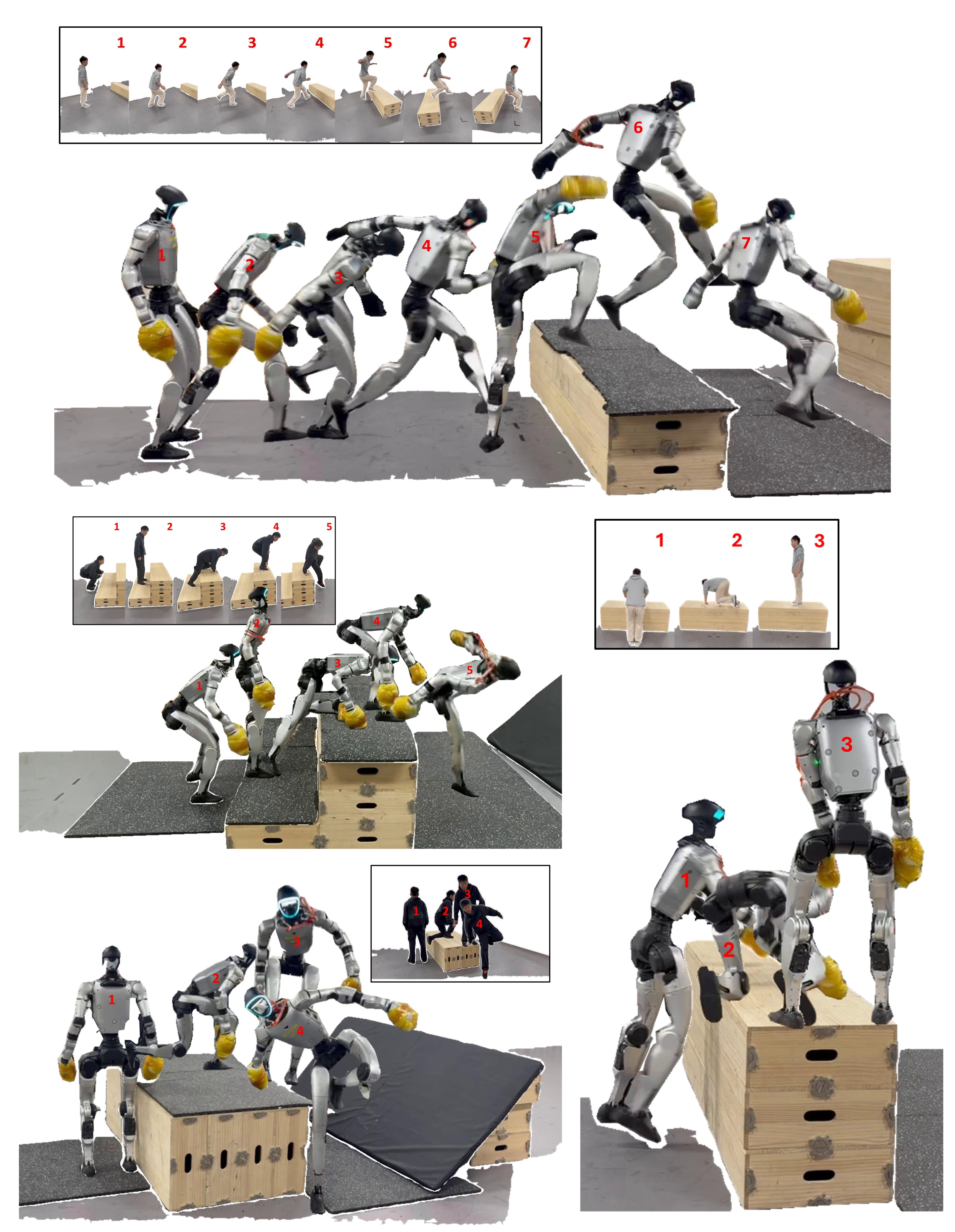}
  \vspace{-0.6em}
  \caption{\textbf{MeshMimic: monocular video-to-humanoid robots.} From ordinary \emph{consumer monocular} videos (\emph{\textbf{no MoCap}}), we reconstruct and optimize human motion together with scene geometry and contacts, then retarget the coupled motion--terrain interactions to humanoid robots that perform dynamic terrain-aware skills.}
  \label{fig:cover_deployment}
  % \vspace{-1.0em}
\end{center} 

\abscontent
\section{Introduction}

Humanoid robots represent one of the most formidable frontiers in robotics. Developing controllers that can manage their high-dimensional degrees of freedom (DoFs) while executing human-like movements remains a profound challenge. With the recent advancements in artificial intelligence, Reinforcement Learning (RL) has emerged as the dominant paradigm for humanoid control, enabling agents to learn complex motor skills through exploratory interaction. However, the sheer complexity of humanoid dynamics makes manual reward engineering and motion design increasingly impractical. Consequently, the field has pivoted toward motion imitation, where robots learn to replicate human behaviors using reference data.

Traditionally, humanoid imitation learning has relied heavily on Motion Capture (MoCap) data \cite{mahmood2019amass}. While high-fidelity, MoCap data is prohibitively expensive to acquire and restricted to controlled laboratory settings. More critically, traditional MoCap often fails to capture the \textbf{geometric context} of the surrounding environment. This decoupling of motion from the physical scene leads to significant physical inconsistencies---such as ``foot skating,'' contact misalignment, or mesh penetration---when the robot is tasked with navigating complex, non-flat terrains. Furthermore, in many real-world scenarios, deploying inertial or optical MoCap systems is logistically impossible. Recent works like \textit{VideoMimic} \cite{videomimic} have attempted to bypass MoCap by using video data; however, they often rely on coarse scene modeling and lack fine-grained contact optimization. Others, such as \textit{OmniRetarget} \cite{omniretarget}, introduce ``Interaction Meshes'' to refine object manipulation but are limited to simple geometric primitives and fail to generalize to irregular, large-scale terrains.

\begin{center}
  \captionsetup{type=figure}
  \includegraphics[width=0.85\textwidth]{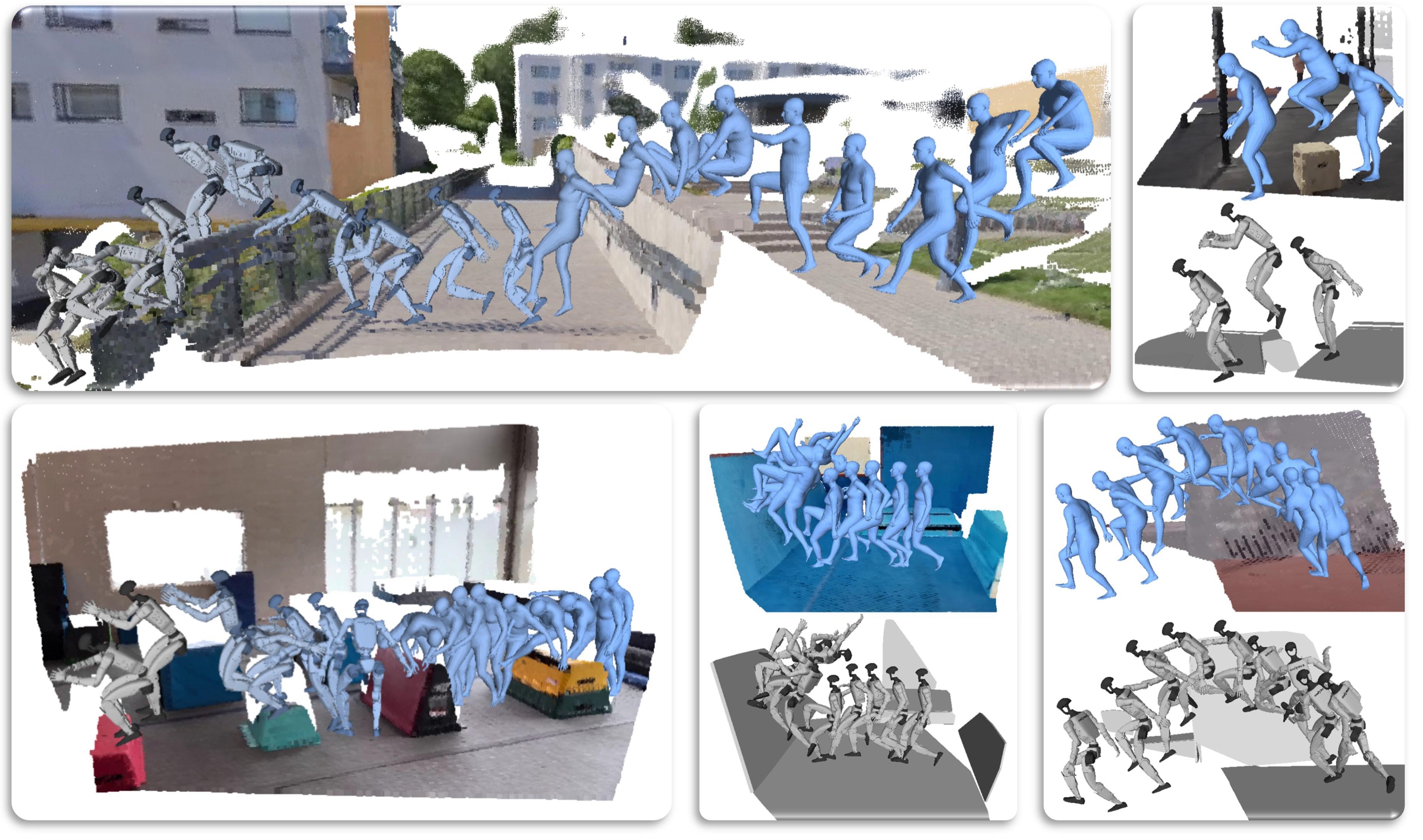}
  \vspace{-0.6em}
  \caption{\textbf{MeshMimic Real-to-Sim.} In-the-wild monocular videos yield \textit{long-horizon motions} over \textit{complex terrains} for contact-consistent \textit{motion--terrain} interaction learning.}
  \label{fig:cover}
  % \vspace{-1.0em}
\end{center}

The rapid evolution of computer vision offers a transformative opportunity to bridge this gap. Modern spatial representations, such as 3D Gaussian Splatting (3DGS) \cite{kerbl20233d} and Neural Radiance Fields (NeRF) \cite{mildenhall2021nerf}, have drastically improved the quality of 3D scene reconstruction. Concurrently, foundation models like SAM3D \cite{SAM3D} have enhanced the robustness of segmentation and 3D generation. Building on these developments, we posit that modern perceptual models can be harnessed to form a unified framework that recovers both high-quality human motion and the associated 3D environmental geometry directly from raw video, enabling downstream RL training.

In this paper, we propose \textbf{MeshMimic}, a novel framework that enables humanoid robots to learn complex, terrain-aware tasks directly from monocular video as shown in Fig.~\ref{fig:cover_deployment}. Our system utilizes 3D vision foundation models to decouple and reconstruct human trajectories and environmental meshes from unposed video sequences. Reconstruction demonstrations are shown in Fig.~\ref{fig:cover}. To handle the inherent noise in visual reconstruction, we introduce a \textbf{Kinematic Consistency Optimization} algorithm to ensure physically plausible reference motions. Furthermore, we present \textbf{MeshRetarget}, a contact-aware retargeting method that maps the intricate interactions between humans and 3D surfaces---such as stepping on uneven rocks or navigating obstacles---onto the humanoid morphology. By integrating these ``physically-grounded'' references into an RL pipeline, our robot learns to perceive and interact with its environment in a unified manner.

% The primary contributions of this work are summarized as follows:
% \begin{itemize}
%     \item \textbf{First Terrain-Aware Humanoid Robot Motion Mimic Framework:} To the best of our knowledge, MeshMimic is the first integrated framework that enables humanoid robots to learn diverse human motor skills across complex 3D terrains by directly perceiving and reconstructing environmental geometry from monocular videos.
%     \item \textbf{A Unified Vision-to-Motion Pipeline:} We introduce an end-to-end pipeline that extracts human motion and 3D scene geometry from consumer-grade monocular video, enabling the learning of complex physical interactions without the need for specialized MoCap hardware.
%     \item \textbf{Kinematic Consistency Optimization:} We develop an optimization strategy to refine noisy 3D visual outputs into high-quality, physically plausible reference trajectories suitable for high-dynamic humanoid RL training.
%     \item \textbf{MeshRetargeting:} We propose a novel retargeting mechanism that prioritizes human-environment interaction features, ensuring that the robot's learned contact states are geometrically consistent with the reconstructed 3D mesh surfaces.
%     \item \textbf{Experimental Validation:} We demonstrate the effectiveness of MeshMimic across a variety of highly dynamic tasks and irregular terrains, showing superior robustness and physical realism compared to existing scene-agnostic baselines.
% \end{itemize}

The primary contributions of this work are summarized as follows:
\begin{itemize}
    \item \textbf{A Terrain-Aware Humanoid Motion Mimic Framework:} We present MeshMimic, an integrated framework that enables humanoid robots to learn diverse motor skills directly from monocular videos. Crucially, we introduce a \textbf{Kinematic Consistency Optimization} strategy during the motion extraction phase, which refines noisy visual pose estimations into physically plausible reference trajectories suitable for control.
    
    \item \textbf{MeshRetarget Mechanism:} We propose a novel retargeting method that explicitly addresses the morphological gap between human subjects and robots. By prioritizing geometric interaction features, MeshRetarget ensures that motions from humans of varying heights are effectively mapped to robots of different dimensions while preserving essential contact constraints.
    
    \item \textbf{Experimental Validation:} We validate our framework across a variety of highly dynamic tasks on irregular terrains. Our experiments demonstrate that MeshMimic achieves superior robustness and physical realism compared to existing scene-agnostic baselines.
\end{itemize}

\section{Preliminaries and Related Works}

\subsection{3D Spatial Modeling and Environment Reconstruction}

The advancement of computer vision has transitioned from sparse point-cloud representations to high-fidelity, dense geometric reconstructions. Early milestones in Structure-from-Motion (SfM) and Multi-View Stereo (MVS), exemplified by frameworks like COLMAP \cite{fisher2021colmap} and \cite{schonberger2016structure}, laid the foundation for spatial mapping but often struggled with textureless surfaces and dynamic occlusions common in real-world robotic environments. The field was further revolutionized by Neural Radiance Fields (NeRF) \cite{mildenhall2021nerf} and its accelerated variants like Instant-NGP \cite{muller2022instant}, which introduced differentiable volumetric representations. More recently, 3D Gaussian Splatting (3DGS) \cite{kerbl20233d} has emerged as a state-of-the-art representation, offering explicit, primitive-based modeling with real-time rendering capabilities.

However, for humanoid robotics, raw geometric reconstruction is insufficient without semantic or instance-level decomposition to distinguish navigable terrain from dynamic agents. While works like LERF \cite{kerr2023lerf} and ConceptFusion \cite{jatavallabhula2023conceptfusion} attempted to ground foundation models into 3D spaces for robotic manipulation, they often lack the fine-grained geometric precision required for high-dynamic locomotion. The emergence of 3D-aware foundation models, such as SAM3D \cite{SAM3D} and Segment Anything in High Quality (HQ-SAM) \cite{ke2023segment}, has enabled robust instance-level segmentation within 3D space. These models allow for the isolation of human actors from their environmental context with unprecedented accuracy. Unlike previous motion imitation works that treat the environment as a simplified static background or a flat plane, our approach leverages these 3D segmentation capabilities to extract the local geometry of the terrain. By converting these segmented instances into high-resolution collision meshes, we provide the necessary physical constraints and exteroceptive observations for downstream reinforcement learning.

\subsection{Humanoid Motion Retargeting}
Motion retargeting is the foundational process of mapping human kinematic trajectories onto a robot's morphology while preserving the semantic and physical intent of the motion. This task is inherently ill-posed due to significant discrepancies in degrees of freedom (DoFs), joint limits, and mass distributions. Early optimization-based approaches, such as GMR \cite{araujo2025retargeting}, focused on preserving geometric relationships and manifold structures to maintain motion fidelity. However, these methods often struggle with the dynamic stability required for high-dimensional humanoid control.

With the rise of large-scale data-driven methods, the field has shifted toward learning-based retargeting and control. Frameworks like PHC \cite{luo2023perpetual} have demonstrated the efficacy of learning robust control policies from large-scale human motion data in simulation. Building upon this, OmniH2O \cite{he2024omnih2o} introduced a universal system for full-body humanoid-to-humanoid and human-to-humanoid mapping, enabling real-time teleoperation and diverse skill acquisition. More recently, Spider \cite{pan2025spider} pushed the boundaries of agile motion retargeting by optimizing for highly dynamic and versatile humanoid behaviors.

Despite these advancements, a critical gap remains in environment-aware retargeting. While OmniRetarget \cite{omniretarget} introduced interaction meshes to refine contacts between the robot and manipulated objects, most current frameworks—including PHC \cite{luo2023perpetual} and OmniH2O \cite{omniretarget} — primarily focus on the agent's internal state or assume a simplified flat-ground plane. This lack of terrain awareness leads to physical inconsistencies, such as "foot skating" or penetration, when the robot interacts with non-planar geometries. Our proposed \textit{MeshRetarget} addresses this by explicitly incorporating high-resolution reconstructed meshes of the terrain into the retargeting loop. By prioritizing contact-invariance on irregular surfaces, we ensure that the retargeted motion is not only kinematically feasible but also geometrically grounded in the actual physical environment.

\subsection{Humanoid Motion Tracking and Whole-Body Control}
Humanoid motion tracking aims to bridge the gap between kinematic reference trajectories and dynamic execution within a physics-based simulator. While early character animation works like DeepMimic \cite{peng2018deepmimic} and ASE \cite{peng2022ase} demonstrated the potential of reinforcement learning (RL) for motion imitation, humanoid robotics requires a higher degree of physical robustness and whole-body coordination (WBC). Recent advancements have shifted toward unified whole-body controllers that can handle the high-dimensional, non-linear dynamics of robotic hardware. 

A significant milestone in this direction is ExBody \cite{cheng2024expressive} and its successor ExBody2 \cite{ji2024exbody2}, which facilitate expressive whole-body control by learning from human motion data. These frameworks emphasize the importance of capturing subtle upper-body gestures alongside stable locomotion, providing a more comprehensive imitation of human behavior than traditional gait-focused controllers. Similarly, frameworks like OmniH2O \cite{he2024omnih2o} have pioneered the "Human-to-Humanoid" pipeline, enabling robots to track diverse human motions in real-time. To scale these capabilities, \textbf{BeyondMimic} \cite{liao2025beyondmimic} and \textbf{VideoMimic} \cite{allshire2025visual} have explored utilizing large-scale datasets and raw video inputs. Similarly, frameworks such as \textbf{Sonic} \cite{luo2025sonic}, \textbf{kungfubot} \cite{xie2025kungfubot} \cite{han2025kungfubot2} and \textbf{UniTracker} \cite{yin2025unitracker} have demonstrated that training on massive motion libraries significantly enhances the generalization of the whole-body controller, allowing for more versatile and robust robotic behaviors across diverse scenarios.

Despite these breakthroughs, a critical limitation persists: most existing humanoid trackers are essentially ``scene-agnostic.'' They typically operate under the assumption of a uniform, flat ground plane and lack exteroceptive awareness of the specific terrain geometry. This prevents the whole-body controller from proactively adjusting its gait or contact points when navigating obstacles—such as stepping over debris or traversing uneven slopes—that were inherent to the original human motion. \textit{MeshMimic} addresses this challenge by integrating high-fidelity 3D scene reconstruction directly into the motion-tracking loop. By grounding the whole-body controller in the reconstructed geometry of the environment, we enable the robot to perform terrain-aware motion imitation that is physically consistent with both the human reference and the environmental constraints.
\section{MeshMimic}
\begin{figure}[t]
  \centering
  \includegraphics[width=0.95\linewidth]{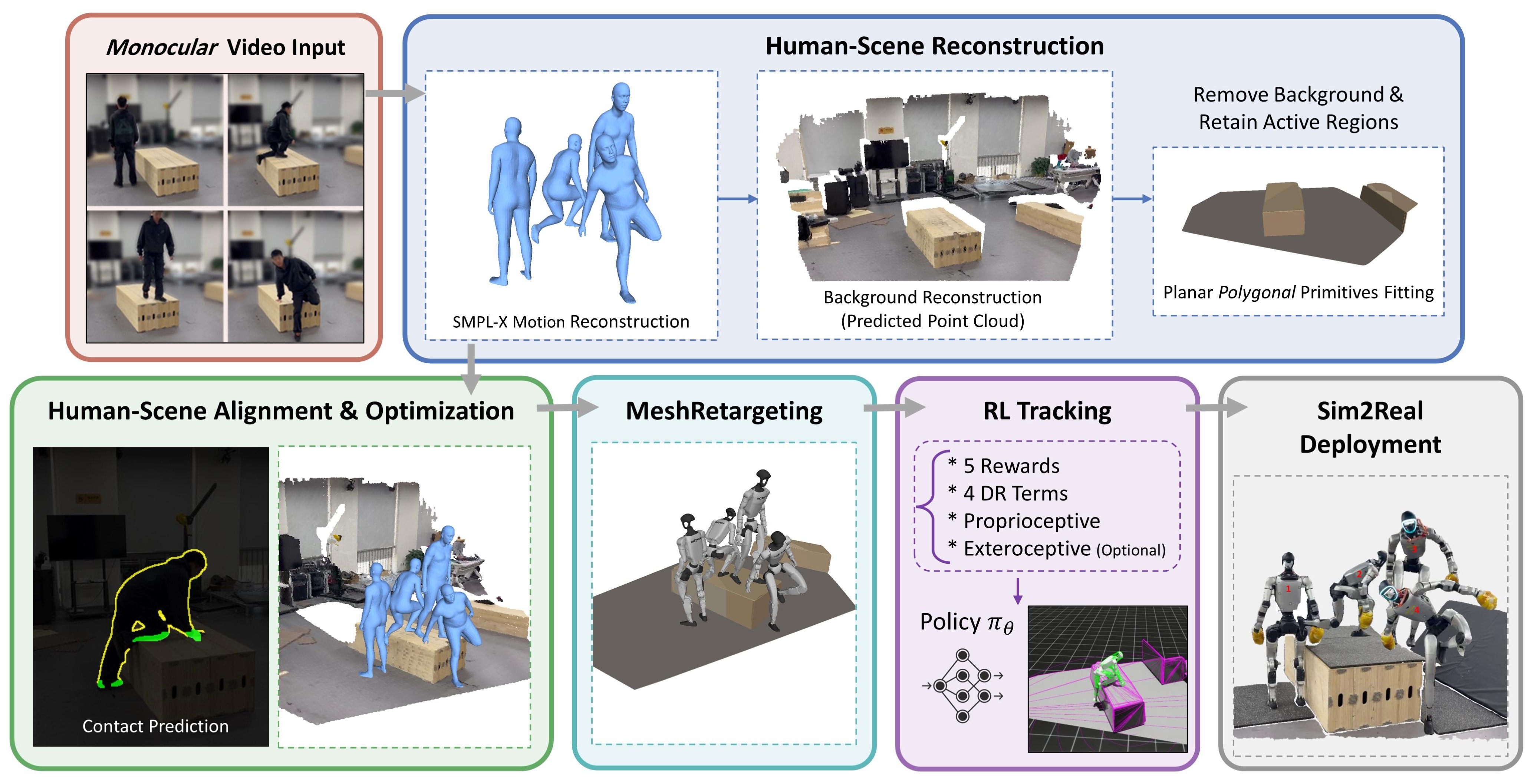}
  \caption{\textbf{MeshMimic Real-Sim-Real Pipeline.} Starting from a monocular video, we reconstruct the scene geometry and human motion, jointly align them to recover metrically consistent human--scene interactions, and retarget the refined motion to a humanoid in simulation for RL policy learning. Finally, we deploy the learned policy to the real robot enabling stable execution over challenging terrain.}
  \label{fig:meshmimic_pipeline}
\end{figure}

The core philosophy of MeshMimic is to bridge the gap between unstructured visual observations and robust humanoid control through a comprehensive \textbf{Real-to-Sim-to-Real} pipeline (Fig.~\ref{fig:meshmimic_pipeline}). This unified framework enables the robot to not only learn from human demonstrations in diverse environments but also to deploy the resulting intelligence back into the physical world.

The \textbf{Real-to-Sim} process constitutes the foundation of our data generation. Starting from a casually captured monocular video, we utilize a 3D-aware perception module to decouple the human actor from the environment. Through a joint optimization of the human trajectory and the reconstructed scene mesh, we recover metrically consistent human-scene interactions (Sec.~\ref{sec:human_scene_recon}). These refined trajectories are subsequently mapped to a humanoid morphology via our \textit{MeshRetargeting} algorithm, which prioritizes contact-invariance and collision avoidance within the reconstructed 3D terrain (Sec.~\ref{sec:retargeting}). This results in a high-fidelity, physically-grounded simulation environment where the humanoid agent can learn complex motor skills via Reinforcement Learning.

The \textbf{Sim-to-Real} phase enables the deployment of learned policies onto physical humanoid robot. By training with terrain-optimized reference trajectories, the whole-body controller (WBC) internalizes the interaction priors necessary for complex environments. This approach ensures that the robot maintains high contact fidelity and physical stability on specific non-flat surfaces, effectively replicating the intricate human-environment interactions captured in the original video.(Sec.~\ref{sec:real_robot})

\subsection{Preprocessing}
\label{sec:preprocessing}
We preprocess a monocular RGB video using off-the-shelf scene reconstruction, detection/tracking, and monocular human reconstruction modules. For the environment, we run $\pi^{3}$~\cite{wang2025pi} to reconstruct the scene and obtain per-frame depth maps $D^{t}$, camera poses $[R^{t}\mid \mathbf{t}^{t}]$, and a shared camera intrinsics $K$. In our scene processing, we depart from approaches that convert point clouds into dense meshes (e.g., VideoMimic) or represent the environment with simple planar primitives (e.g., CRISP~\cite{CRISP}). Instead, we approximate the scene using planar polygonal primitives. This representation effectively suppresses noisy points in dynamic reconstructions, provides a simple yet faithful scene description, and captures richer geometric structure than conventional planar primitives. For the human, we detect the target person using ViTDet~\cite{ViTDet} and associate the identity across frames via SAM2~\cite{SAM2}. Given the tracked person instances, we reconstruct per-frame human body geometry and motion using SAM3D~\cite{SAM3D}. Specifically, we follow the official SAM3D-Body pipeline: we convert the intermediate MHR representation to SMPL-X~\cite{SMPL-X}, yielding per-frame local pose parameters $\boldsymbol{\theta}^{t}$, body shape $\boldsymbol{\beta}$, and 3D SMPL joints $\mathbf{J}^{t}_{\text{3D}} \in \mathbb{R}^{J\times 3}$, together with the estimated orientation $\mathbf{\phi}^{t}$ translation $\mathbf{t}^{t}$ in camera coordinate.

Importantly, both the scene reconstruction (camera parameters and geometry from $\pi^{3}$) and the monocular human reconstruction are not metrically scaled. Moreover, the recovered SMPL-X motion is expressed in the camera coordinate system and thus is not directly comparable across frames in a common world frame. In the subsequent stage, we therefore jointly optimize the human motion and scene geometry to recover a metrically consistent, world-aligned human trajectory and environment geometry.

\subsection{Human-Scene Reconstruction}
\label{sec:human_scene_recon}
Unlike VideoMimic, which focuses on curated videos with clean motion, our Internet-crawled clips often exhibit rapid camera/subject motion, leading to jitter, blur, and occlusions. In this regime, learning-based contact prediction like BSTRO~\cite{bstro} becomes unstable and degrades human--scene optimization. We address this with depth-edge--guided contact prediction, metric-scale human--scene alignment and joint human-scene optimization for robust contacts and metrically consistent reconstruction.

\begin{figure}[t]
  \centering
\includegraphics[width=0.95\linewidth]{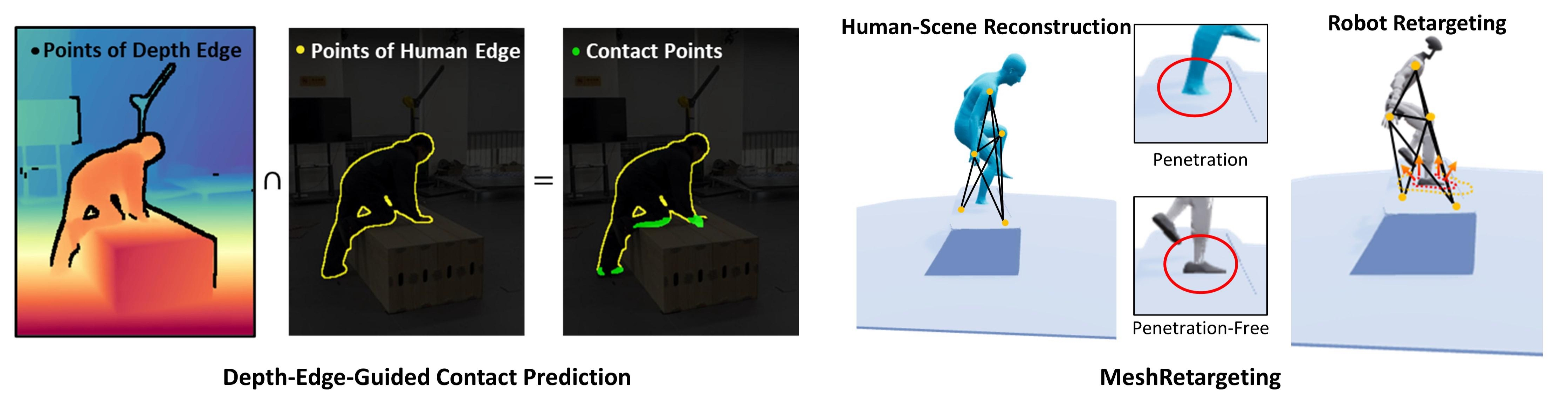}
\caption{\textbf{Left:} Depth-edge--guided contact prediction. \textbf{Right:} MeshRetargeting optimization for penetration correction and contact-consistent retargeting.}

  \label{fig:contact}
\end{figure}
\textbf{Depth-edge–guided contact prediction}. Visual cues such as depth edges $E_{\text{depth}}$ from monocular depth and silhouette boundaries $E_{\text{human}}$ from human segmentation are relatively stable. We exploit them to extract reliable human--scene contacts. We compute $E_{\text{human}}$ by applying a morphological gradient to the binary human mask, and dilate $E_{\text{depth}}$ to form an exclusion region $\tilde{E}_{\text{depth}}$ around depth discontinuities.We then define the contact band as the silhouette-boundary pixels not covered by $\tilde{E}_{\text{depth}}$, i.e., $\mathcal{P}_c=\{\,p\in\mathcal{P}_{\text{human}}\mid E_{\text{human}}(p)=1 \wedge \tilde{E}_{\text{depth}}(p)=0\,\}$, where $p=(u,v)$ denotes a pixel, $\mathcal{P}_{\text{human}}$ is the set of silhouette-boundary pixels, and $E_{\text{human}}(p),\tilde{E}_{\text{depth}}(p)\in\{0,1\}$ indicate whether $p$ lies on the human boundary or within the dilated depth-edge exclusion region, respectively. We further dilate $\mathcal{P}_c$ with a small kernel to improve robustness to projection noise.
In Figure.~\ref{fig:contact}.A, human-scene contacts are consistently highlighted in green across the video sequence. Finally, background points whose projections fall within this band are selected as candidate scene contacts.
\label{contact_prediction}

\textbf{Metric-scale human–scene alignment}. We jointly optimize the human trajectory and the scene scale. Since SAM-3D Body provides a strong initialization, we keep the SMPL-X pose and shape fixed and optimize only global alignment variables: the per-frame translations $\mathbf{t}^{0:T}$ and a single scene scale $\alpha$ applied to the reconstructed point cloud. Using the metric height prior of SMPL-X as a reference, $\alpha$ corrects the scale mismatch between scene reconstruction and human-derived metric scale. Let $\mathbf{H}$ denote the SMPL-X function that outputs vertices $\mathcal{V}_h^t=\mathbf{H}(\boldsymbol{\beta},\boldsymbol{\theta}^t,\mathbf{t}^t,\boldsymbol{\phi}^t)$. 
Following~\cite{physic}, we only optimize the global translation $\mathbf{t}^t$ using an alignment loss $L_{\text{align}} = L_{J2d} + L_d$. Here, $L_{J2d}$ measures the 2D joint reprojection error between SMPL-X--regressed joints (projected with the camera intrinsics) and SAM3D-body 2D keypoints prediction, while $L_d$ is a symmetric Chamfer distance between camera-facing SMPL-X vertices and the metric-scale human point set. To avoid matching backside vertices to the human points, we compute $L_d$ using only camera facing vertices, selected by the angle between each vertex normal and viewing direction to be less than $65^\circ$.

\textbf{Kinematic consistency optimization.} 
Although the human vertices $\mathcal{V}_h^t$ and metric-scale scene points provide a strong initialization for human--scene alignment, occlusions, inaccurate depth estimates, and noisy 2D keypoint predictions can still cause interpenetration or hovering artifacts, as well as unstable and drifting trajectories. To improve physical plausibility, we further enforce human--scene kinematic consistency by introducing additional constraints. Specifically, we jointly optimize the per-frame translations $\mathbf{t}^{0:T}$ and a single global scene scale using an alignment term together with contact, penetration, trajectory smoothness, and foot-snapping regularization:
\begin{equation}
L_{\text{total}}=\lambda_{align}L_{align}+\lambda_{c}L_{c}+\lambda_{p}L_{p}+\lambda_{sm}L_{sm}+\lambda_{\text{fs}}L_{\text{fs}}.
\label{eq:total_loss}
\end{equation}
We detail the contact loss ($L_c$), penetration loss ($L_p$), trajectory smoothness loss ($L_{sm}$), and foot-snapping loss ($L_{fs}$) in the following. 

% \textcolor{red}{The weighting coefficients $\lambda_*$ are provided in the supplementary material.}

\noindent\textit{Contact loss ($L_c$).}
We encourage predicted contacting human vertices to coincide with the estimated scene contact locations.
For each frame $t$, we obtain a set of scene contact points $\{\mathbf{c}_j^t\}$ and the corresponding human vertex indices $\{i_j^t\}$.
Since the reconstructed scene is re-scaled by a single global factor $\alpha$, we enforce contact consistency by matching each scaled scene contact point $\alpha\,\mathbf{c}_j^t$  to its corresponding human vertex $\mathbf{v}_{h,i_j^t}^t \in \mathcal{V}_h^t$:
\begin{equation}
\label{eq:contact_loss}
L_c=\frac{1}{\sum_t |\mathcal{C}^t|}\sum_t \sum_{j\in \mathcal{C}^t}
\left\| \alpha\,\mathbf{c}_j^t- \mathbf{v}_{h,i_j^t}^t \right\|_2^2,
\end{equation}
which anchors contact vertices to the scene and reduces hovering at predicted support regions.

\textit{Penetration loss ($L_p$).} To discourage the human mesh from intersecting scene geometry, we construct a TSDF volume~\cite{volumetric, kinectfusion} from the background point cloud and oriented normals.
We then query the TSDF at all world-space human vertices using trilinear sampling.
Denote the sampled signed distance at vertex $\mathbf{v}$ by $d(\mathbf{v})$, where $d>0$ is outside and $d<0$ indicates penetration.
We introduce a slack $\tau$ and only penalize penetration deeper than $-\tau$, and denote penalty as  $p(\mathbf{v})=\max\!\bigl(0,\;-(d(\mathbf{v})+\tau)\bigr)$.
Finally, we apply a Huber-style robust penalty to stabilize gradients:
\begin{equation}
L_p \;=\; \frac{1}{|\mathcal{V}|}\sum_{\mathbf{v}\in\mathcal{V}}
\mathrm{Huber}\!\left(p(\mathbf{v})\right).
\label{eq:penetration_loss}
\end{equation}
so shallow violations are softly corrected while deeper intersections are strongly penalized.

\textit{Trajectory smoothness loss ($L_{sm}$).} To mitigate frame-to-frame jitter and drift in the recovered global motion, we regularize the per-frame global translation trajectory.
Let $\mathbf{T}^t=\mathbf{t}_{\text{cam}}^t+\mathbf{t}^t$ denote the global translation at frame $t$, and let $N$ be the number of frames in the sequence ($t=0,\ldots,N-1$). Inspired by~\cite{prompthmr}, we penalize both velocity and acceleration using finite differences scaled by the frame rate $f$:
\begin{equation}
L_{sm} \;=\; \frac{1}{N-1}\sum_{t=0}^{N-2}\left\|(\mathbf{T}^{t+1}-\mathbf{T}^{t})\,f\right\|_2^2
\;+\;
\frac{1}{N-2}\sum_{t=0}^{N-3}\left\|(\mathbf{T}^{t+2}-2\mathbf{T}^{t+1}+\mathbf{T}^{t})\,f\right\|_2,
\end{equation}
which encourages temporally coherent motion while still allowing genuine fast movements.

\textit{Foot-snapping loss ($L_{fs}$).} To reduce near-ground foot hovering (i.e., ``foot snap'' artifacts), we explicitly encourage foot joints to lie on the scene surface when they are already close to it.
We extract foot joint positions $\mathbf{q}_{f}^t$ from the optimized 3D keypoints, evaluate their TSDF values $d(\mathbf{q}_{f}^t)$, and activate the term only within a narrow near-surface band $\mathbb{I}\!\left(0<d(\mathbf{q}_{f}^t)\le \tau_{\text{contact}}\right)$, where $\tau_{\text{contact}}$ is a contact threshold. Within this band, we penalize squared distances to pull the feet onto the surface:
\begin{equation}
L_{fs}\;=\;\frac{1}{N}\sum_{t,f}
\mathbb{I}\!\left(0<d(\mathbf{q}_{f}^t)\le \tau_{\text{contact}}\right)\, d(\mathbf{q}_{f}^t)^2.
\end{equation}
These two terms $L_{fs}$ and $L_p$ are complementary: $L_{fs}$ reduces near-surface hovering by penalizing small \emph{positive} TSDF distances, whereas $L_p$ discourages interpenetration by penalizing \emph{negative} TSDF distances.

% \begin{figure}[t]
%   \centering
% \includegraphics[width=0.6\linewidth]{figures/meshretargeting.jpeg}
%  \caption{\textbf{MeshRetargeting.}}
%   \label{fig:meshretargeting}
% \end{figure}

\subsection{MeshRetargeting}
\label{sec:retargeting}
Following OminiRetarget~\cite{omniretarget}, we preserve the spatial relationships among robot parts, manipulated objects, and terrain by minimizing the Laplacian deformation energy of an interaction mesh built from corresponding human/robot anatomical keypoints together with sampled object and terrain points. In large-scale scenes, the selection of sampled terrain points is critical. If the sampled terrain points are far from the human, the Laplacian deformation energy may change only marginally even when the local retargeting quality is poor. To address this, we sample not only global terrain points but also additional points in the vicinity of the human. This strategy preserves global geometric proportions while improving local alignment accuracy. We solve for the robot configuration $q_t$ per frame with an SQP-style optimizer, under hard constraints for collision avoidance, joint/velocity limits, and stance-foot anchoring to prevent foot skating.  

During retargeting, the human motion may remain collision-free while the retargeted robot still penetrates the terrain due to kinematic mismatch. To improve physical plausibility, we apply a lightweight TSDF-based correction to the robot \emph{global translation}. 
Specifically, we build a TSDF of the terrain and query the signed distance $d(\mathbf{x})$ at the robot vertices $\mathcal{V}_r^t$, where penetration is indicated by $d(\mathbf{v})<0$.
Following the feasibility criterion in Eq.~\ref{eq:penetration_loss}, we seek an offset $\Delta\mathbf{o}$ such that $\min_{\mathbf{v}\in\mathcal{V}_r^t} d(\mathbf{v}+\Delta\mathbf{o}) \ge \tau_{\textit{safety}}$, where $\tau_{\textit{safety}}<0$ allows a small tolerance.
To make the update well-defined, we first compute a unit correction direction $\mathbf{u}$ from the average SDF gradient over penetrated (or near-surface) vertices,
\begin{equation}
\mathbf{u}=\frac{\frac{1}{|\mathcal{M}|}\sum_{\mathbf{v}\in\mathcal{M}} \nabla d(\mathbf{v})}
{\left\|\frac{1}{|\mathcal{M}|}\sum_{\mathbf{v}\in\mathcal{M}} \nabla d(\mathbf{v})\right\|_2},
\end{equation}
and then parameterize the offset as $\Delta\mathbf{o}=\eta\,\mathbf{u}$.
We choose the smallest $\eta\ge 0$ via line search such that the feasibility constraint holds, yielding a collision-free global translation.
\section{Experiments}
\subsection{Reconstruction Comparison}

\paragraph{Evaluation Setup.}
We evaluate the robustness of our reconstruction pipeline on a subset of the SLOPER4D dataset~\cite{sloper4d}. Following established evaluation protocols~\cite{videomimic,wham,tram}, we report performance on two complementary aspects: (i) \,\emph{human trajectory reconstruction} and (ii) \,\emph{scene geometry reconstruction}. For benchmarking, we use a subset of SLOPER4D that includes only sequences where SAM2 tracking---comprising human detection and cross-frame association---is successful. This subset contains two sequences for each activity category: running, walking, and stair ascent/descent.

\paragraph{Metrics and Baselines.}
For human trajectories, we report World-frame Mean Per Joint Position Error (W-MPJPE) and World-frame Aligned MPJPE (WA-MPJPE). For each sequence, we partition the motion into 100-frame segments. W-MPJPE aligns only the first two frames of each segment to the ground truth, emphasizing global consistency over long horizons, while WA-MPJPE aligns the entire segment and measures local trajectory accuracy over time.

For scene geometry, we report the Chamfer Distance (in meters) between the aligned predicted point cloud and the LiDAR point cloud, restricted to the RGB camera's field of view.

We compare against WHAM~\cite{wham}, TRAM~\cite{tram}, and VideoMimic~\cite{videomimic}. WHAM focuses on human motion reconstruction and does not recover the environment (thus Chamfer Distance is not applicable). For fair comparison, all baselines are reproduced using their official implementations.

\paragraph{Results.}
As summarized in Table~\ref{tab:recon_comp}, our method achieves the best overall performance on both human motion and scene reconstruction. Compared with VideoMimic, we reduce WA-MPJPE from 112.13 to 94.32 (\,\(-15.9\%\)\,) and W-MPJPE from 696.62 to 518.98 (\,\(-25.5\%\)\,), indicating improved local fidelity and substantially better global trajectory stability. For scene geometry, our Chamfer Distance decreases from 0.75 to 0.61 (\,\(-18.7\%\)\,), suggesting more accurate and less noisy terrain reconstruction.

Compared with TRAM, our improvements are more pronounced, especially for geometry: Chamfer Distance drops from 10.66 to 0.61, highlighting the benefit of explicitly modeling and reconstructing the surrounding environment. Overall, these gains validate that our reconstruction pipeline provides higher-quality human--scene inputs, which are critical for downstream contact reasoning and physics-based humanoid learning.

\begin{figure}[t]
  \centering
\includegraphics[width=0.95\linewidth]{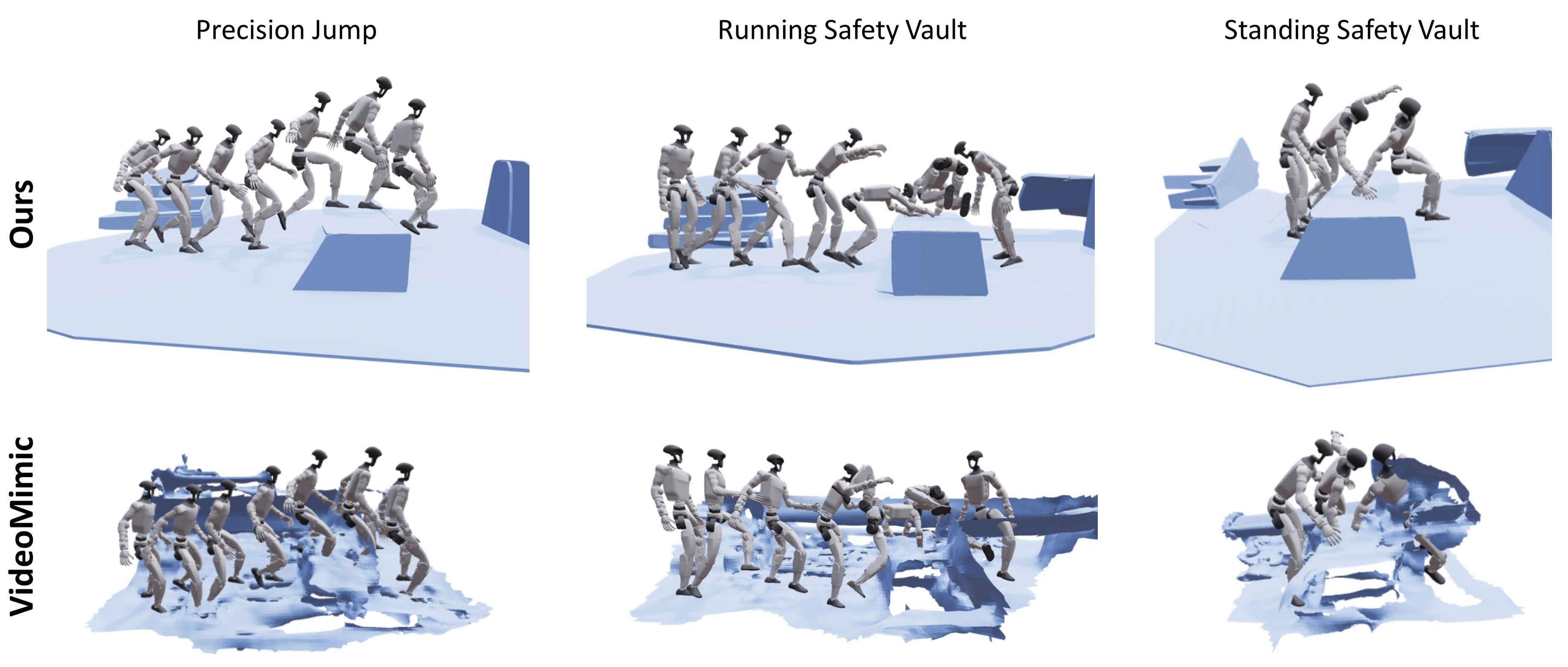}
 \caption{\textbf{Comparison with VideoMimic.}}
  \label{fig:contact}
\end{figure}

\subsection{Training and Deployment Configuration}
\label{sec:real_robot}

\begin{table}[t]
    \centering
    \small
    \setlength{\tabcolsep}{6pt}
    \begin{tabular}{lccc}
        \toprule
        Methods & WA-MPJPE & W-MPJPE & Chamfer Distance \\
        \midrule
        WHAM$^\ast$~\cite{wham} & 189.29 & 1148.49 & -- \\
        TRAM~\cite{tram}        & 149.48 & 954.90  & 10.66 \\
        VideoMimic~\cite{videomimic} & 112.13 & 696.62 & 0.75 \\
        \midrule
        \textbf{Ours} & \textbf{94.32} & \textbf{518.98} & \textbf{0.61} \\
        
        \bottomrule
    \end{tabular}
    \caption{\textbf{Comparison of Reconstruction.} $\ast$~WHAM does not recover the environment.}
    \label{tab:recon_comp}
\end{table}

Our method leverages high-fidelity human--scene reconstruction to obtain both high-quality kinematic reference motions and watertight scene geometry. Such accurate reconstruction significantly reduces the burden on reward shaping: rather than relying on elaborate humanoid RL reward engineering~\cite{li2025reinforcement, zhao2025resmimic, weng2025hdmi}, we adopt a minimal BeyondMimic-style formulation~\cite{liao2025beyondmimic}. As summarized in Table~\ref{tab:train_config}, our configuration uses only generic tracking terms and standard regularization, yet it is sufficient for robust whole-body tracking with contact-rich scene interactions. Beyond the BeyondMimic-style observation design, we additionally incorporate a global torso position signal during training; during real-robot deployment, we obtain torso position from an optical motion-capture system. We train the policy in IsaacLab~\cite{mittal2025isaac} using asymmetric PPO, where the actor operates on proprioceptive and reference features while the critic has access to privileged scene. Unlike prior interactive humanoid controllers that introduce contact- or task-specific reward heuristics, we find that high-quality reconstruction and reference motions already provide strong implicit supervision.

\begin{table}[h!]
\centering
\small
\begin{tabularx}{\linewidth}{l X}
\toprule
\textbf{Category} & \textbf{Description} \\
\midrule
\textbf{Actor Observation} 
& Reference Motion: joint position/velocity, future steps torso position/orientation error \\
& Proprioception: projected gravity, torso angular velocity, joint position/velocity all with histories \\
& Previous Action: executed policy action with histories \\
\midrule
\textbf{Critic Observation} 
& Reference Motion: joint position/velocity, future steps torso position/orientation error \\
& Proprioception: projected gravity, torso linear/angular velocity, body links position/orientation, joint position/velocity with histories \\
& Previous Action: executed policy action with histories \\
\midrule
\textbf{Reward}
& Anchor Tracking: anchor position/velocity/orientation error \\
& Body Tracking: tracking term for body position, orientation, linear and angular velocity \\
& Action Rate: penalize rapid action change \\
& Soft Joint Limit: penalize limit violation \\
\midrule
\textbf{Termination}
& Large anchor position tracking deviation on Z-axis \\
& Large anchor orientation tracking deviation \\
& Large body position tracking deviation on Z-axis \\
\midrule
\textbf{Domain Randomization}
& Randomize robot body material \\
& Torso COM: $\pm 0.025$m (x), $\pm 0.05$m (y), $\pm 0.05$m (z) \\
& Joint default position: $\pm 0.01$ rad \\
& Random push: $0.3$ m/s, $0.78$ rad/s for $(1$--$3)$ s \\
\midrule
\textbf{Motion Adaptive Sampling}
& BeyondMimic-style motion bin adaptive sampling \\
\bottomrule
\end{tabularx}
\vspace{-4pt}
\caption{Training configuration.}
\label{tab:train_config}
\end{table}

Training interactive imitation directly from scratch is computationally demanding due to slower simulation caused by complex contact computation. To improve efficiency, we first pre-train a generic whole-body motion tracker~\cite{chen2025gmt, luo2025sonic} on $\sim$50 hours of non-interactive human motion data using the same asymmetric PPO setup. We then fine-tune the policy on scene-interactive motion references. In practice, we adopt lightweight feed-forward architectures for asymmetric PPO: both actor and critic are 4-layer MLPs with hidden dimensions \([3072, 1536, 768, 512]\) and operate with 5-step observation histories and a 5-step future motion horizon. After fine-tuning on scene-interactive data, the resulting policy runs onboard a Unitree~G1 robot at 50\,Hz using an NVIDIA Jetson Orin. Because the reconstruction pipeline preserves exact real-world terrain geometry, we deploy the robot within the same real-world scenes used for video capture.

\subsection{Real2Sim2Real Comparison}

\paragraph{Scene Setup.}
We evaluate our real2sim2real pipeline across eight diverse scene-interaction tasks that involve stepping, vaulting, climbing, and parkour-like contact patterns. Specifically, the testing scenes are:

\begin{itemize}[noitemsep,topsep=2pt]
    \item \textbf{walk1}: walk on a flat plane.
    \item \textbf{jump box1 (JB1)}: jump onto a 40\,cm box.
    \item \textbf{jump box2 (JB2)}: running single-leg jump onto a 40\,cm box and drop down.
    \item \textbf{climb box1 (CB1)}: climb onto a 50\,cm box, walk to edge, and descend using single-hand support.
    \item \textbf{climb box2 (CB2)}: side climb onto a 60\,cm box.
    \item \textbf{safety vault1 (SV1)}: single-hand safety vault over a 40\,cm box.
    \item \textbf{safety vault2 (SV2)}: double-hand safety vault over a 40\,cm box.
    \item \textbf{jump climb down1 (JCD1)}: jump onto a 20\,cm box, climb onto a 60\,cm box, and descend using single-hand support.
    % \item \textbf{jump step down jump1 (JSDJ1)}: jump onto a 40\,cm box, walk up to 60\,cm box, descend using single-hand support, and perform single-leg jump cycles.
    % \item \textbf{jump step down vault1 (JSDV1)}: jump onto a 40\,cm box, walk up to 60\,cm box, descend using single-hand support, and double-hand vault.
\end{itemize}

(Please refer to the supplementary video for visualizations.)

\paragraph{Baselines and Metrics.}

We primarily compare against \textbf{VideoMimic}~\cite{videomimic}, which similarly provides a monocular real2sim2real pipeline. The comparison is conducted at two levels: (i) \emph{real2sim} via the mean training reward in IsaacLab, and (ii) \emph{sim2real} via real-world success rate (SR). The mean rewards follows the formulation in Table~\ref{tab:train_config} and is measured after 40k PPO iterations with 2048 parallel IsaacLab environments. SR is computed over 10 real-world trials per scene, where a trial is counted as successful if the robot completes the full motion sequence without intervention.

\begin{figure*}[t]
    \centering
    \begin{subfigure}[t]{0.49\linewidth}
        \centering
        \includegraphics[width=\linewidth]{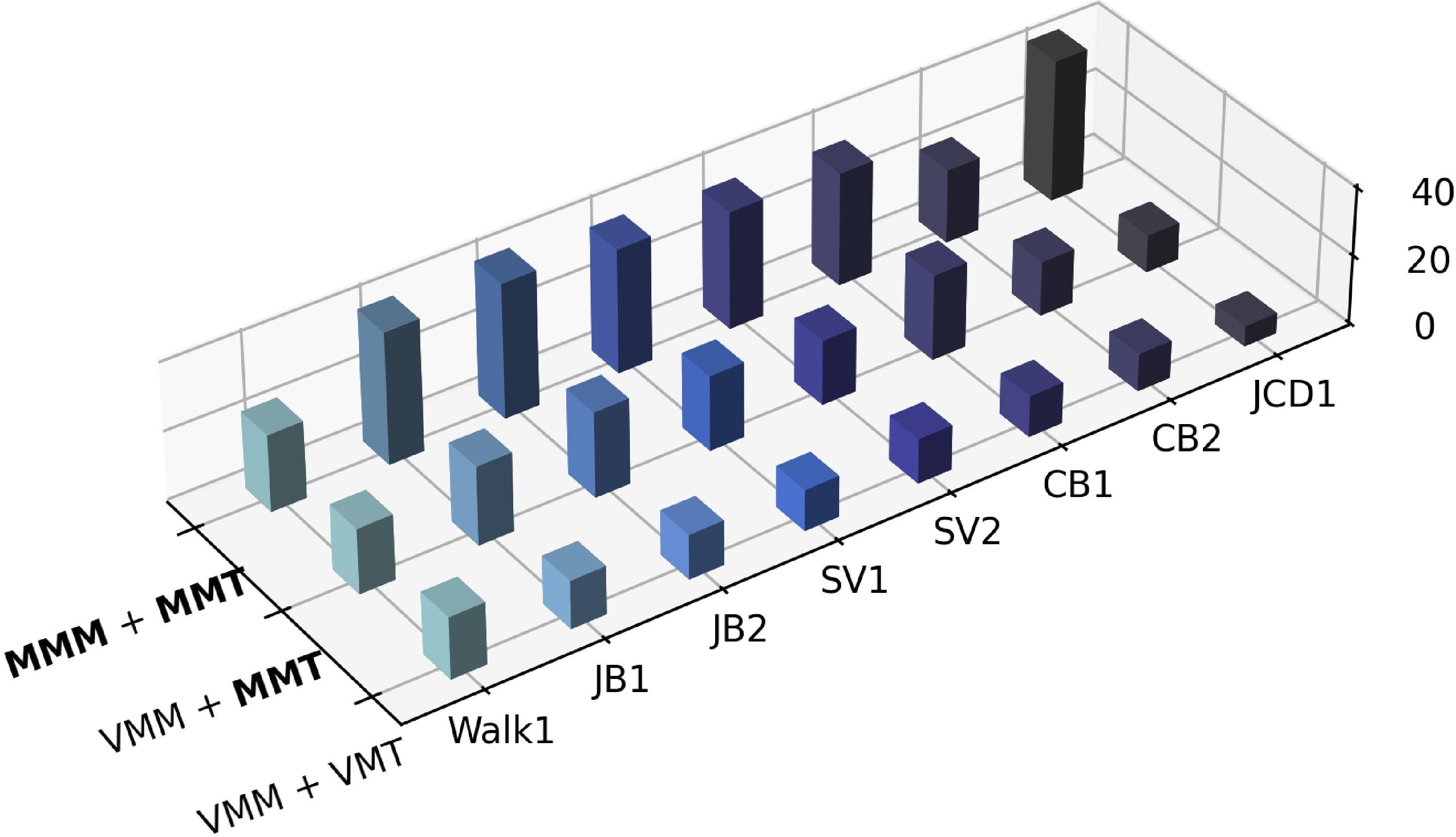}
        \caption{Training Rewards}
        \label{fig:real2sim2real_reward}
    \end{subfigure}
    \hfill
    \begin{subfigure}[t]{0.49\linewidth}
        \centering
        \includegraphics[width=\linewidth]{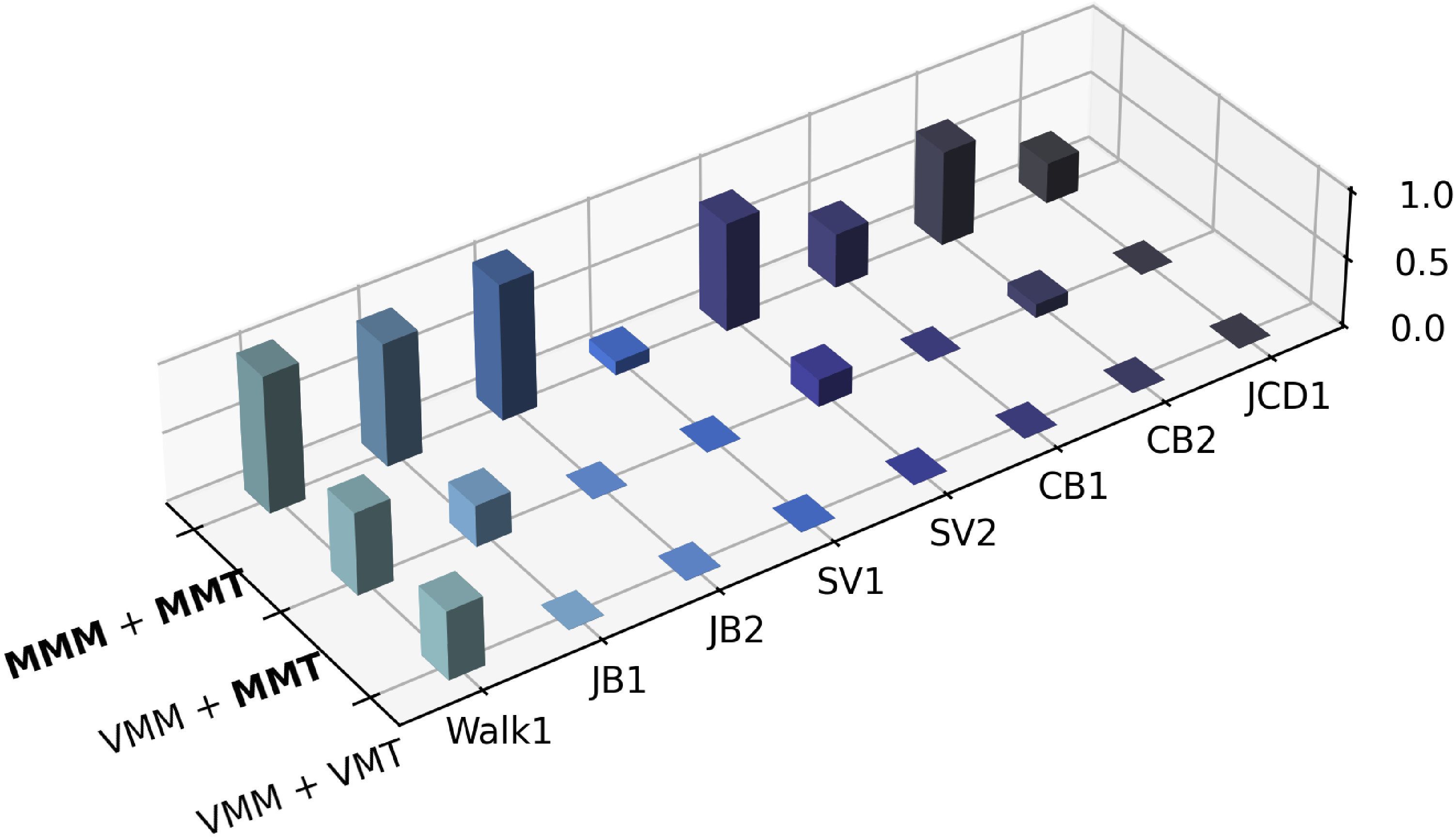}
        \caption{Deployment Success Rate}
        \label{fig:real2sim2real_sr}
    \end{subfigure}
    \caption{Effect of motion and terrain reconstruction on training and deployment performance (\textbf{MMM: MeshMimic Motion}; VMM: VideoMimic Motion; \textbf{MMT: MeshMimic Terrain}; VMT: VideoMimic Terrain).}

    \label{fig:real2sim2real}
\end{figure*}

\paragraph{Terrain and Motion Reconstruction.}
As shown in Fig.~\ref{fig:contact}, VideoMimic struggles to reconstruct terrain geometry due to background clutter and limited camera viewpoints, leading to blurred edges, floating or hollow surfaces, and uneven ground topology. Such artifacts significantly complicate contact reasoning during training and deployment. To decouple motion and terrain effects, we evaluate reconstruction in three configurations: \textbf{MMM+MMT} (MeshMimic motion + MeshMimic terrain), \textbf{VMM+MMT} (VideoMimic motion + MeshMimic terrain), and \textbf{VMM+VMT} (VideoMimic motion + VideoMimic terrain). Quantitative results are shown in Fig.~\ref{fig:real2sim2real}.

In terms of \emph{Training reward} (Fig.~\ref{fig:real2sim2real_reward}), VideoMimic motions exhibit frequent foot-in-air artifacts, interpenetrations, and frame-to-frame drift. These errors make it difficult for the policy to track under physical constraints, especially in long-horizon or contact-intensive scenes. As a result, MMM+MMT consistently achieves higher mean rewards than VMM+MMT, with the largest gaps observed in long or multi-contact scenes such as JCD1. Using VMM+VMT further reduces reward due to inaccurate terrain geometry: floating obstacles block humanoid trajectories, hollow geometry causes unexpected falls during contact, and irregular surfaces interfere with adaptive initialization during training. These effects collectively increase motion-tracking difficulty during training.

For \emph{real-world deployment} (Fig.~\ref{fig:real2sim2real_sr}), MMM+MMT achieves stable success rates across most tasks, with failures mainly occurring in highly dynamic settings (e.g., SV1) or long-horizon multi-scene interactions (e.g., CB1, JCD1). In contrast, VMM+MMT fails sim2sim validation (IsaacLab $\rightarrow$ MuJoCo) on several tasks (JB2, SV1, CB1, JCD1), preventing safe deployment. Even for tasks that pass sim2sim checks (Walk1, JB1, SV2, CB2), deployment remains unstable due to motion drift inherent in VideoMimic reconstructions. With VMT, sim2sim failures become more severe (JB1, JB2, SV1, SV2, CB1, CB2, JCD1), again preventing deployment except in Walk1, where terrain interaction is minimal.

Overall, \textbf{MMM+MMT} yields the highest simulation reward and the highest real-world success rate. We attribute this to accurate reconstruction of both motion and terrain, which reduces model mismatch throughout the real2sim2real loop and enables reliable contact reasoning in challenging parkour-like scenes.

\paragraph{Global Torso Position as an Observation.}
We further investigate whether exposing the policy to global torso position improves sim-to-real deployment. This signal is unavailable from onboard proprioception and must be externally estimated during deployment (e.g., via optical motion capture), thus serving as an exteroceptive cue.

As shown in Table~\ref{tab:pos_ablation}, injecting global torso position significantly benefits long-horizon traversal behaviors that require sustained locomotion and multi-contact interactions across obstacles. In particular, we observe success rate improvements of $+20\%$ on \texttt{JB2}, $+20\%$ on \texttt{CB1}, and $+30\%$ on \texttt{JCD1}. These motions span meters and involve transitions between stepping, jumping, and climbing, during which small drift in global pose accumulates into meter-scale deviation. Access to global position mitigates this drift, improving foot placement accuracy and reducing failure cases where the robot reaches an obstacle ``off-phase'' or misaligns with the terrain.

\begin{table}[t]
\centering
\begin{tabular}{lcccccccc}
\toprule
         & Walk1 & JB1 & JB2 & SV1 & SV2 & CB1 & CB2 & JCD1 \\
\midrule
w/o pos   & 1.0 & 1.0 & 0.8 & 0.7 & 1.0 & 0.2 & 1.0 & 0.0 \\
w/ pos    & 1.0 & 0.9 & 1.0 & 0.5 & 0.8 & 0.4 & 0.7 & 0.3 \\
\bottomrule
\end{tabular}
\caption{Effect of adding global torso position to the observation space on sim-to-real deployment success rate.}
\label{tab:pos_ablation}
\end{table}

Conversely, short-duration yet highly dynamic motions (e.g., \texttt{SV1}, \texttt{SV2}, \texttt{CB2}) show degraded performance, with decreases of $-20\%$, $-20\%$, and $-30\%$ respectively. We observed two contributing factors: (1) fast upper-body and contact transitions introduce intermittent marker occlusions during motion capture, amplifying noise in global position estimates, and (2) these tasks rely more heavily on local agility and contact timing than on accurate global pose. In such regimes, noisy global inputs may destabilize tracking behavior and hinder recovery from mis-steps.

Taken together, these results suggest that global observability is beneficial for long-horizon and path-dependent behaviors, while purely proprioceptive policies remain preferable for short, high-acceleration motions where perception noise dominates the benefit of improved global alignment.

\section{Conclusion and Future Work}

We present a humanoid motion learning framework built purely on RGB video data: without any motion-capture system, markers, or specialized sensing hardware, we directly reconstruct both human motions and the surrounding terrain from videos and transfer them into agile parkour-style skills with robust tracking. This ``in-the-wild'' data pipeline makes the approach readily applicable to arbitrary outdoor scenarios and significantly lowers the cost of data collection compared to traditional MoCap-based solutions. Through extensive experiments, our method demonstrates stable performance across diverse motions and disturbances, indicating strong potential for deployment on real robotic platforms. As future work, we will push toward vision-based generalized parkour over diverse, previously unseen terrains, and develop a fully closed-loop system that tightly couples perception, planning, and control for long-horizon navigation.

% \section*{Acknowledgements}

\clearpage
% \appendix

% \begin{center}
% {\LARGE \textbf{Supplementary Material}}\\[0.5em]
% \end{center}
% \vspace{1em}

% \input{sections/supplementary}

\clearpage
\setcitestyle{numbers}
\bibliographystyle{plainnat}
\bibliography{main}

\end{document}